\documentclass{article}

\PassOptionsToPackage{numbers, compress}{natbib} 
\usepackage[final]{nips_2017}

\usepackage[utf8]{inputenc} 
\usepackage[T1]{fontenc} 
\usepackage{hyperref} 
\usepackage{url} 
\usepackage{booktabs} 
\usepackage{amsfonts} 
\usepackage{nicefrac} 
\usepackage{microtype} 

\usepackage{graphicx}
\usepackage{multirow}
\usepackage{tabularx}
\usepackage{makecell}
\usepackage{color}
\usepackage{textcomp}

\title{NSML: A Machine Learning Platform That Enables You to Focus on Your Models}

\author{
  Nako Sung\textsuperscript{1}, Minkyu Kim\textsuperscript{1}, Hyunwoo Jo\textsuperscript{1}, Youngil Yang\textsuperscript{1},\\
  \textbf{Jinwoong Kim\textsuperscript{1}, Leonard Lausen\textsuperscript{1,4}, Youngkwan Kim\textsuperscript{1}},\\ \textbf{Gayoung Lee}\textsuperscript{3}
  \textbf{Donghyun Kwak\textsuperscript{2}, Jung-Woo Ha\textsuperscript{1}, and Sunghun Kim\textsuperscript{1,4}}\\
  Clova AI Research, NAVER Corp., Seongnam, Korea\textsuperscript{1}\\
  Search Solution Inc.\textsuperscript{2}, NAVER WEBTOON Corp. \textsuperscript{3}\\
  Hong Kong University of Science Technology\textsuperscript{4}\\
  \texttt{\{nako.sung, min.kyu.kim, hyunwoo.jo, youngil.yang,} \\
  \texttt{jinwoong.kim, leonard.lausen, ykkim.kim\}@navercorp.com}\\
  \texttt{gayoung.lee@webtoonscorp.com}\\
  \texttt{\{donghyun.kwak, jungwoo.ha, sung.kim.n\}@navercorp.com} }

\begin{document}

\maketitle

\begin{abstract}
  Machine learning libraries such as TensorFlow and PyTorch simplify
  model implementation. However, researchers are still required to perform a
  non-trivial amount of manual tasks such as GPU allocation, training status
  tracking, and comparison of models with different hyperparameter settings. We propose a system to handle these tasks and help researchers focus
  on models. We present the requirements of the system based on a collection of
  discussions from an online study group comprising 25k members. These include automatic GPU allocation, learning status visualization, handling model parameter snapshots as well as hyperparameter modification during learning, and comparison of performance metrics between
  models via a leaderboard. We describe the system architecture that fulfills
  these requirements and present a proof-of-concept implementation, NAVER Smart  Machine Learning (NSML). We test the system and confirm substantial efficiency  improvements for model development.
\end{abstract}

\section{Introduction}

Deep learning \cite{lecun2015deep} has recently enabled remarkable progress in 
diverse fields including as speech \cite{arik2017deep, amodei2016deep},
computer vision \cite{huang2016densely, redmon2016yolo9000}, machine translation
\cite{gehring2017convolutional}, and natural language processing
\cite{lin2017structured}. In addition, multiple easy-to-use deep learning
frameworks such as TensorFlow \cite{abadi2016tensorflow}, PyTorch
\cite{chintala2017overview} and MXNet \cite{chen2015mxnet} significantly reduce the barrier to implementing such deep
learning models.

Still, performing experiments on deep learning architectures requires
considerable time and computational resources, and becomes only feasible by
leveraging multiple servers and GPUs. These requirements introduce a non-trivial
amount of supporting tasks into a researchers' workflow, such as GPU allocation
and distribution of code and data to servers. As a result, researchers often
spend large portions of their time on such incidental tasks.

Next to computational requirements, hyperparameter sensitivity of many deep
learning models forces researchers to run their experiments using different
hyperparameter combinations and keep track of the results, often in a manual
fashion. To debug the training process and get better understanding of
hyperparameter effects visualization packages such as TensorBoard
\cite{zou2017understanding} and Visdom
\footnote{https://github.com/facebookresearch/visdom} can be used, but support
to compare concurrently performed experiments on many models is lacking. Some
machine learning platforms such as Google Cloud ML platform
\footnote{https://cloud.google.com/products/machine-learning/} aim to reduce
these issues. However, vendor lock-in and support of a restricted set of deep
learning libraries only pose their own challenges.

In this paper, we first collect requirements for enhancing research efficiency,
and then describe a system architecture that fulfills the requirements. Based on
the architectue design, we implement NAVER
\footnote{https://www.navercorp.com/en/index.nhn} Smart Machine Learning (NSML).
The requirements cover many incidental subtasks from automatic GPU allocation
and releases to a leaderboard for comparing the model performances, which are
not core but essential tasks conducted manually by many researchers. NSML
automatically conducts these subsidiary subtasks instead of humans and provides
web-based interfaces that are easy to use, thus enabling researchers to focus on
their model implementation and data. Currently, NSML supports diverse deep
learning libraries with Python APIs including TensorFlow, PyTorch, and MXNet,
which are the most popular in deep learning research communities. In addition, NSML
supports not only servers with single GPU but also GPU-clusters.

We implemented NSML on a server cluster equipped with 80 P40 GPUs as a
prototype. We evaluate NSML based on alpha tests conducted by researchers at
NAVER. They implemented models for three tasks: face emotion recognition, face
generation, and movie rating prediction. Comments from NSML users indicate that
NSML dramatically enhances the research efficiency.
 
The contributions of NSML are summarized as follows:
\begin{itemize}
\item{We define the requirements for efficient deep learning research.}
\item{We designed and implemented a new deep learning research platform that
    enhances the research efficiency.}
\item{We demonstrate the usages of NSML on three real-world applications.}
\end{itemize}

\section{Common Workflow Challenges in Machine Learning Research}

We list challenges observed from an online community
\footnote{https://www.facebook.com/groups/TensorFlowKR} comprising more than
25,000 global deep learning researchers and developers over a time frame of one
year.

\noindent\textbf{Difficulty in resource management:} With increasingly deep and
complex models, increasing computational requirements come along. However, it is
often difficult to acquire sufficient resources for training models. For example,
to train ResNet-152 \cite{he2016deep}, eight GPUs are required to optimize the
152 layers in a timely fashion. Consider a case where the total number of GPUs
in a cluster is sufficient, but due to bad scheduling no single server with
eight idling GPUs is available, so that the model cannot be trained. It is
therefore necessary to appropriately allocate each task to the server based on
the amount of resources required. However, manual assignment is likely to cause
inefficiency if multiple developers share the server.

\noindent\textbf{Difficulty in data management:} Deep learning methods require large amounts of training data. Popular datasets
such as ImageNet \cite{russakovsky2015imagenet} or YouTube-8M
\cite{abu2016youtube} have sizes of multiple gigabytes or more. Researchers must
ensure that the datasets are available on each server on which their models
should be trained, which so far has to be performed manually. Furthermore, it is inefficient to access
the dataset every time separately or to duplicate the preprocessing for
running multiple experiments with the same dataset.

\noindent\textbf{Lack of integrated management tools:} Researchers need to
perform multiple experiments to compare the performances between different
models and to find better hyperparameters. These experiments often run
simultaneously on multiple servers in order to reduce training time. To manage
and control each training task, it is so far necessary for researchers to manually access
each of the servers for collecting and comparing results. These tasks are
time-consuming and could be automatized.

\noindent\textbf{Difficulty in tracking experiment environments over time:} Over
time researchers often change model structure or experiment hyperparameters in
the quest for improved model performance. Without tedious and manual recordings,
it can be difficult to reproduce past experiments as the information about the
exact conditions under which an experiment was conducted may be lost.

\section{NSML: NAVER Smart Machine Learning}
\subsection{Requirements of ML Platforms}
We elicit the following requirements for practical large-scale ML systems based
on the challenges identified in the previous section and the comments from the
deep learning community.

\noindent\textbf{Resource Management}
\begin{itemize}
\item{Better computational resource management to improve utilization and job
    scheduling.}
\item{Automatic execution of the entire machine learning pipeline, beginning
    with data preprocessing to finding the best model, whenever e.g. the dataset
    is updated.}
\end{itemize}

\noindent\textbf{Data Management}
\begin{itemize}
\item{Users should be able to post datasets once and reuse them for multiple
    models.}
\item{Users should be able to share datasets with others.}
\item{The systems should be capable of preprocessing large datasets.}
\end{itemize}

\noindent\textbf{Serverless Configuration}
\begin{itemize}
\item{The systems should not require any configuration or (manual) installation
    from users.}
\item{The systems should not be limited to the use of specific libraries and
    should not depend on specific platforms.}
\item{The servers should be hidden from users, and users should be able to use
    the system in a serverless manner by just submitting a training or testing
    tasks to the platform.}
\end{itemize}

\noindent\textbf{Experiment Management and Visualization}
\begin{itemize}
\item{The systems should handle parallel runs with different jobs priorities.}
\item{The systems should automatically visualize and summarize the learning
    progress.}
\end{itemize}

\noindent\textbf{Leaderboard}
\begin{itemize}
\item{The systems should provide a leaderboard which can compare various models and
    hyperparameters for each dataset.}
\end{itemize}

\noindent\textbf{AutoML}
\begin{itemize}
\item{The systems should be able to predict the performance of experiments based on
    previously run experiments.}
\item{The systems should be able to automatically optimize the hyperparameters
    based on the performance predictions.}
\item{The systems should save the model of best score.}
\end{itemize}


\subsection{NSML System Architecture}

\begin{figure}
  \begin{center}
  \includegraphics[width=0.8\columnwidth]{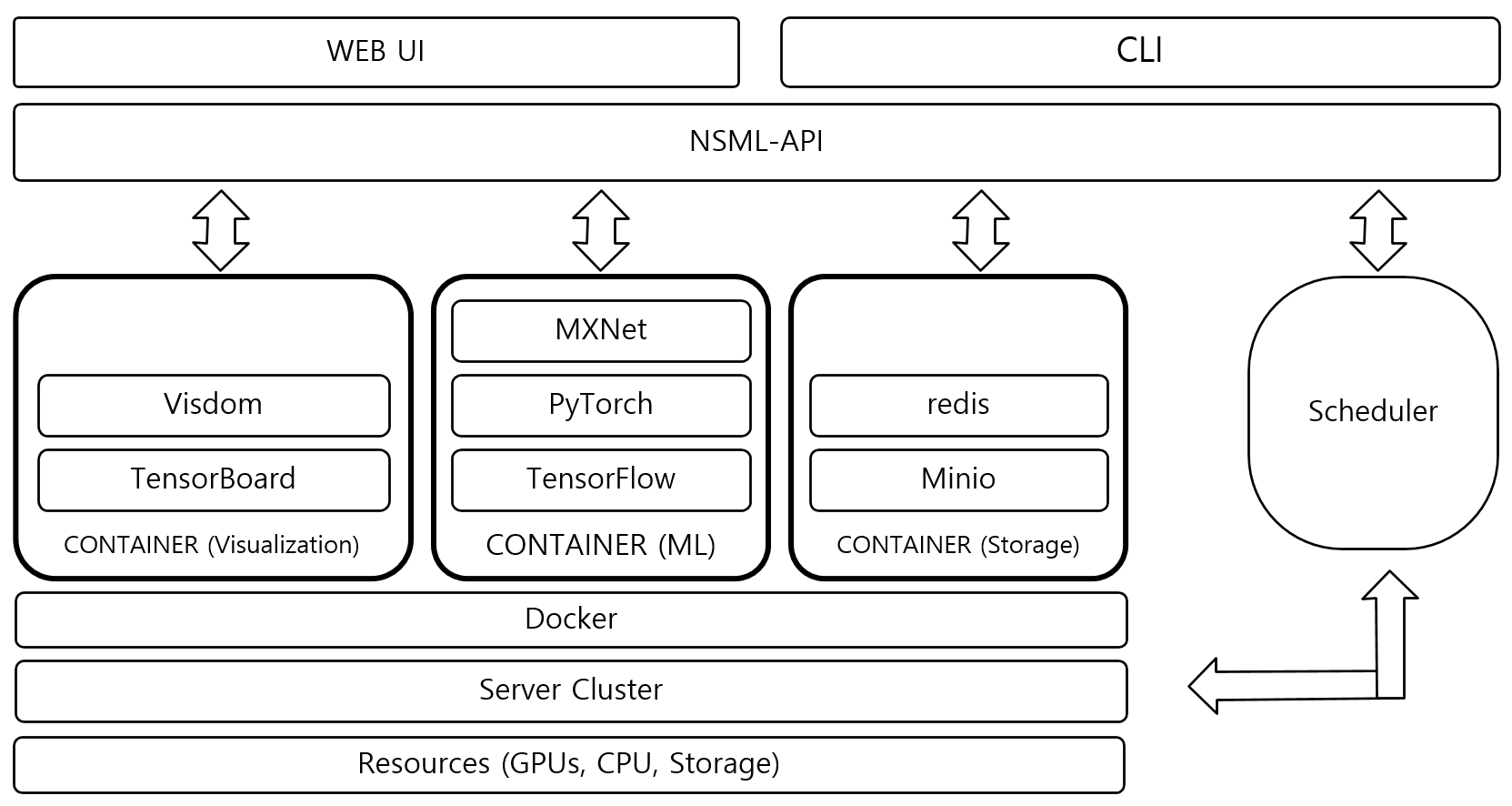}
  \caption{System Architecture of NSML}
  \label{fig:system}
  \end{center}
\end{figure}
We described the system architecture of NSML as shown in Figure
\ref{fig:system}(a) in detail. NSML consists of three main parts: scheduler,
containerized system, and user interface (UI).

\noindent\textbf{Scheduler:}
NSML has a scheduler with a centralized scheduling model based on the
master-slave structure for efficient allocation of computational resources such
as CPUs and GPUs. In NSML, a single node (i.e., master node) is in charge of
monitoring all computational resources and scheduling tasks for all clients.
Other nodes (i.e., slave nodes) collect information about their computational
resources and periodically report it to the master node. A centralized model
often suffers from a single point of failure (SPOF) \cite{armbrust2010view}. We
handle this issue with the leader election process by electing new master node
as in \textit{Zookeeper} \cite{hunt2010zookeeper} when the master node fails. To
train models, clients have to submit a job to the scheduler for obtaining
computational resources. Then, the scheduler first looks over the job queue. If
the job queue is empty, the scheduler immediately selects an available slave
node and informs the client about its address. Compared to first inserting the
job into the queue, this approach allows the scheduler to avoid queue operation
overhead. Otherwise, the job will be inserted in the job queue.

\noindent\textbf{Containerized System:}
We separate containerized systems into two groups: storage containers and ML
containers. Storage containers use \textit{minio}
\footnote{https://www.minio.io/} to store and supply datasets to ML containers.
They also store the performance of all models trained with the respectively
provided dataset as well as display the results in a leaderboard to make clear
which model performed best. Storage containers furthermore back up intermediate
and final results of trained models and also store the source code associated
with the experiments so that users can easily reproduce and improve both their
own and other users’ models. ML containers can contain any type of systems and
libraries. When a user sets up an environment, NSML automatically packages it
into a ML container and copies the user’s codes and datasets from the
respective storage containers. Then NSML runs the code, reports the learning
status and backs up intermediate and final results of trained models.

Containerized ML system can solve many challenges that frequently appear during
training deep learning models. For instance thanks to the containerized
infrastructure different users can rely on different library versions for their
machine learning environment via NSML. If one user wants to use PyTorch in
python 2.7, he or she just needs to select the corresponding base docker image.
Meanwhile other users can use TensorFlow in python 3.6 and their models can be
run on the same machine thanks to the isolation provided by the containers. NSML
also provides visualization of trained model via both TensorBoard and Visdom
using the ML containers to simplify model evaluation.

\noindent\textbf{User Interface:}
The user interface (UI) controls the input and output flow between users and
NSML. The UI is separated into the command line interface \textit{NSML-CLI} and
the \textit{web UI}. \textit{NSML-CLI} runs on users' local system and requires
docker to prepare containers for NSML. \textit{NSML-CLI} collects the model code
in a container. Usage of \textit{NSML-CLI} may be challenging for users that are
not familiar with the command line. The \textit{web UI} wraps \textit{NSML-CLI}
in a web application and is more intuitive and easy to use for these users.
Furthermore the \textit{web UI} provides visualizations such as graphs, logs,
and demos.

\subsection{Implementation Details}
NSML is a decentralized cloud system except the scheduler. A user directly
connects to a docker container via CLI once the container is scheduled.
\textit{NSML-CLI} is implemented in Python 3.

Research on machine learning consists of three main steps: preparation and
preprocessing data, training models and testing models.

NSML supports data preparation and preprocessing steps with web storage such as
\textit{minio}. \textit{NSML-CLI} communicates with a storage container via
\textit{docker-cli} and Python libraries.

For training models, we first implemented a scheduler to distribute physical
resources including GPUs efficiently. After allocation of resources, NSML builds
a docker image for ML environments and mounts a dataset from a storage
container. These two steps are bottlenecks of training models in NSML. We
removed the first bottleneck by reusing existing docker images if a user needs
the same environment. The other can be solved by sharing dataset directories
among all ML containers when they are physically located at the same host
machine. Then NSML runs trains the model by running the user provided code.
During training NSML stores intermediate trained models into the storage
container. With these backup files, NSML supports reproducing the same model and
tuning hyperparameters during training. We implemented both hyperparameter
tuning in training time and model evaluation through python read-eval-print loop
(REPL), which allows a user to evaluate one simple line and print the result.
With this feature, NSML can achieve hyperparameter tuning in training time by
pausing user-written codes, downloading a model from storage container, and
resuming the code. We used this feature not only to test and resume models, but
also to demonstrate models as a service through \textit{web UI}.

Our web servers run with \textit{nginx} \footnote{https://nginx.org/en/} and the
\textit{web UI} translates a user’s input to \textit{NSML-CLI} commands.
For example the \textit{web UI} allows testing trained models in real time by
issuing the \texttt{nsml infer} command. For this command \textit{nsml-CLI}
creates new container with python REPL, and a user can run a demo of trained
models through this new container.

\begin{figure}
  \begin{center}
  \includegraphics[width=0.8\columnwidth]{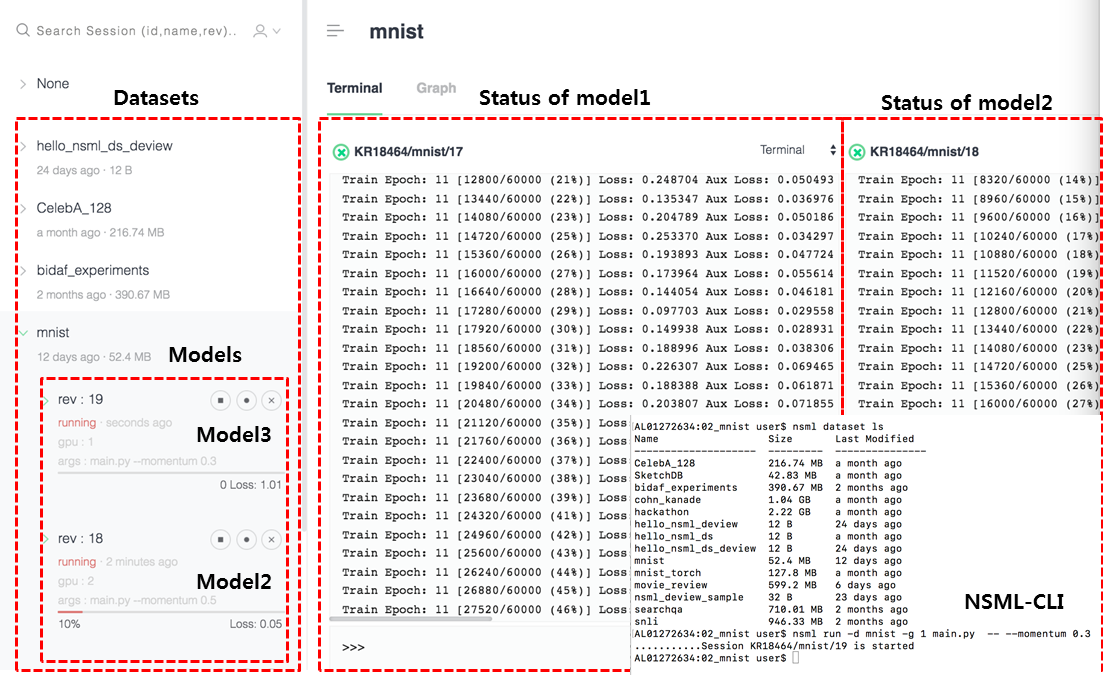}
  \caption{Example of Web UI and NSML-CLI}
  \label{fig:web-cli}
  \end{center}
  \vskip -0.1in
\end{figure}

\subsection{Usage Examples}
After installation, \texttt{nsml} commands can be invoked with \texttt{nsml
  [OPTIONS] COMMAND [ARGS]...} syntax. NSML has two main commands: \texttt{nsml
  dataset} and \texttt{nsml run}. \\  
\texttt{nsml dataset} provides commands to
examine and manage the list of datasets available in the NSML cloud environment
as well as to add new datasets to the environment. \\
\texttt{nsml run} provides functionality to pack codes from the local workstation of the researcher and execute it in the cloud environment with access to a specified dataset that will be mounted at a predefined mounting point.

For example, \texttt{nsml run main.py -d [name of dataset]} would package the
code in the current directory, send it to the NSML server, and then run
\texttt{main.py} giving access to the specified dataset.

To ease the task of performing experiments, a client package that can be
accessed from the experiment code is provided. It contains functions that allow
experiment metrics to be logged via both TensorBoard and Visdom. These logs can be
accessed in raw form with \texttt{nsml logs [OPTIONS] SESSION} and
visualized \texttt{nsml plot [OPTIONS] SESSION}, where \texttt{SESSION} denotes
a session identifier returned by \texttt{nsml run}.

Furthermore, NSML contains a kaggle-like leaderboard on which experiment results
on a specific dataset can be recorded and compared. \texttt{nsml dataset board [OPTIONS] DATASET} command gives access to the NSML leaderboard. Figure \ref{fig:web-cli} display an example of \textit{Web UI} and \textit{NSML-CLI} on the MNIST dataset.

\section{Experiments via Alpha Tests}
\subsection{Details of alpha tests}
\begin{figure}[ht]
  \begin{center}
  \includegraphics[width=1.0\columnwidth]{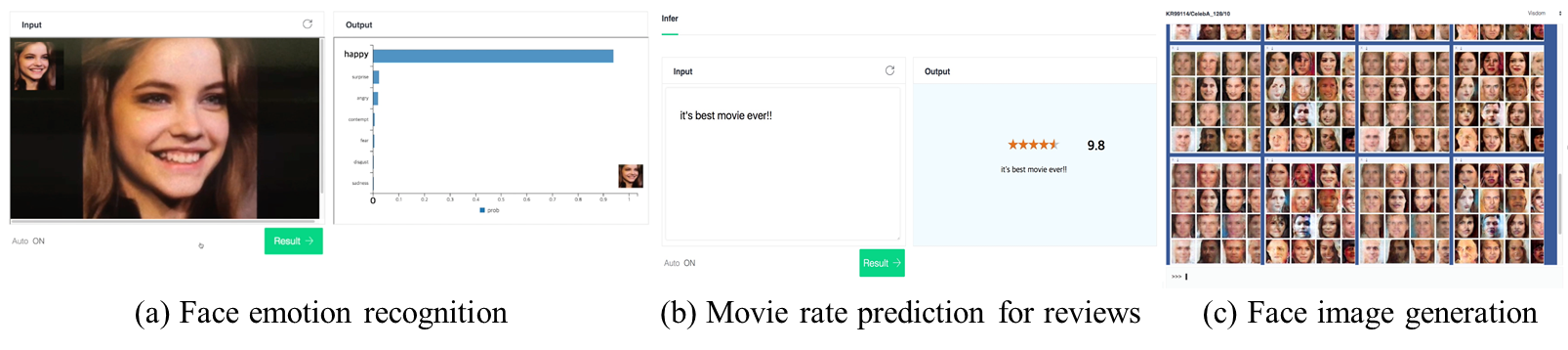}
  \caption{Visualization of three real-world tasks}
  \label{fig:learn-inf}
  \end{center}
\end{figure}

\begin{figure}[ht]
  \begin{center}
  \includegraphics[width=1.0\columnwidth]{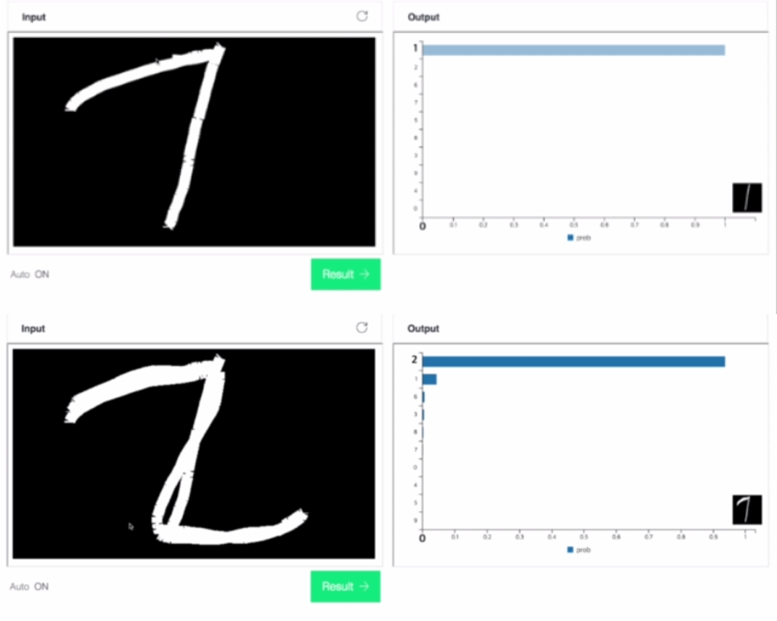}
  \caption{Demonstration of immediate classification on interactive user input using a trained model from the MNIST dataset. Input in the lower figure was modified from the upper figure by adding some lines, and this causes the class label probability to be changed from 1 to 2.}
  \label{fig:learn-mnist}
  \end{center}
\end{figure}

We operated NSML in a phase of alpha tests with four different machine learning tasks including 1) MNIST classification, 2) GAN-based face generation, 3) BiLSTM-based movie rate prediction, and 4) CNN-based facial emotion recognition. Figure \ref{fig:learn-inf} illustrates the demonstration of the experiments on three datasets. As shown in Figure \ref{fig:learn-inf}, users can conveniently develop their model and perform the experiments.In addition, Figure~\ref{fig:learn-mnist} depicts a web demonstration of real-time classification on the input a user draws using the model trained from the MNIST dataset. NSML supports interactive web demonstration using trained models, thus enabling researchers to easily visualize the results on user input and output of their models as shown in Figure~\ref{fig:learn-mnist}.   

\subsection{Interviews of alpha testers}
During the alpha tests, we collected reviews from the alpha testers and summarized them into following items.
\begin{itemize}
\item{The usage of NSML is more intuitive than other platforms. It is not hard to learn and understand the work flow of the system. And the NSML provides almost all the necessary functions such as multiple GPUs allocation and reproducing the past experiments. It makes it easier to focus on researching and developing the deep learning models.}
\item{The visualization and history logging of experiments make it more efficient. We don't need to manually implement tedious jobs such as drawing a graph and saving a model.}
\item{Currently, the system is little bit unstable. Sometimes the system has no response and has been recovered after a few minutes. Maybe the system is not ready for public services.}
\end{itemize}

\section{Discussion and Concluding Remarks}
We demonstrate a new machine learning platform that enhance the research efficiency, NSML. For designing NSML, we investigate the challenges in machine learning research from a deep learning community. Then, we define the detailed requirements with six categories. NSML provide the functionalities corresponding to the requirements, thus enabling the researchers to focus on their model implementation and data. 

Current version of NSML has one main limitations. NSML only supports data-driven machine learning tasks. Machine learning tasks such as reinforcement learning require an environment for agents as well as a dataset. This issue will be addressed by treating an environment in a way similar to that NSML treats a dataset. More AutoML features will be added in future for dramatically reducing research time. Ultimately, we add a protocol that the best model for each task can be automatically copied to A/B test systems for NAVER and Clova\footnote{https://clova.ai/en/research/research-areas.html} services in order to seamlessly connect between research and services. 


More detailed information will be available at https://clova.ai/en/research/research-area-detail.html?id=1. 

\subsubsection*{Acknowledgments}
The authors thank Hyunah Kim for data preparation, Doris Choi for discussion, and Miru Ryu for the web interface design. In particular, the authors appreciate all members of TensorFlow KR in Facebook for discussing diverse requirements. 






\bibliographystyle{IEEEbib.bst} \bibliography{nsml}

\begin{thebibliography}{10}

\bibitem{lecun2015deep}
Yann LeCun, Yoshua Bengio, and Geoffrey Hinton,
\newblock ``Deep learning,''
\newblock {\em Nature}, vol. 521, no. 7553, pp. 436--444, 2015.

\bibitem{arik2017deep}
Sercan Arik, Gregory Diamos, Andrew Gibiansky, John Miller, Kainan Peng, Wei
  Ping, Jonathan Raiman, and Yanqi Zhou,
\newblock ``Deep voice 2: Multi-speaker neural text-to-speech,''
\newblock {\em arXiv preprint arXiv:1705.08947}, 2017.

\bibitem{amodei2016deep}
Dario Amodei, Sundaram Ananthanarayanan, Rishita Anubhai, Jingliang Bai, Eric
  Battenberg, Carl Case, Jared Casper, Bryan Catanzaro, Qiang Cheng, Guoliang
  Chen, et~al.,
\newblock ``Deep speech 2: End-to-end speech recognition in english and
  mandarin,''
\newblock in {\em International Conference on Machine Learning}, 2016, pp.
  173--182.

\bibitem{huang2016densely}
Gao Huang, Zhuang Liu, Kilian~Q Weinberger, and Laurens van~der Maaten,
\newblock ``Densely connected convolutional networks,''
\newblock {\em arXiv preprint arXiv:1608.06993}, 2016.

\bibitem{redmon2016yolo9000}
Joseph Redmon and Ali Farhadi,
\newblock ``Yolo9000: better, faster, stronger,''
\newblock {\em arXiv preprint arXiv:1612.08242}, 2016.

\bibitem{gehring2017convolutional}
Jonas Gehring, Michael Auli, David Grangier, Denis Yarats, and Yann~N Dauphin,
\newblock ``Convolutional sequence to sequence learning,''
\newblock {\em arXiv preprint arXiv:1705.03122}, 2017.

\bibitem{lin2017structured}
Zhouhan Lin, Minwei Feng, Cicero Nogueira~dos Santos, Mo~Yu, Bing Xiang, Bowen
  Zhou, and Yoshua Bengio,
\newblock ``A structured self-attentive sentence embedding,''
\newblock {\em arXiv preprint arXiv:1703.03130}, 2017.

\bibitem{abadi2016tensorflow}
Mart{\'\i}n Abadi, Ashish Agarwal, Paul Barham, Eugene Brevdo, Zhifeng Chen,
  Craig Citro, Greg~S Corrado, Andy Davis, Jeffrey Dean, Matthieu Devin,
  et~al.,
\newblock ``Tensorflow: Large-scale machine learning on heterogeneous
  distributed systems,''
\newblock {\em arXiv preprint arXiv:1603.04467}, 2016.

\bibitem{chintala2017overview}
Soumith Chintala,
\newblock ``An overview of deep learning frameworks and an introduction to
  pytorch,''
\newblock 2017.

\bibitem{chen2015mxnet}
Tianqi Chen, Mu~Li, Yutian Li, Min Lin, Naiyan Wang, Minjie Wang, Tianjun Xiao,
  Bing Xu, Chiyuan Zhang, and Zheng Zhang,
\newblock ``Mxnet: A flexible and efficient machine learning library for
  heterogeneous distributed systems,''
\newblock {\em arXiv preprint arXiv:1512.01274}, 2015.

\bibitem{zou2017understanding}
Xuan Zou,
\newblock ``Understanding deep learning through visualization,''
\newblock 2017.

\bibitem{he2016deep}
Kaiming He, Xiangyu Zhang, Shaoqing Ren, and Jian Sun,
\newblock ``Deep residual learning for image recognition,''
\newblock in {\em Proceedings of the IEEE conference on computer vision and
  pattern recognition}, 2016, pp. 770--778.

\bibitem{russakovsky2015imagenet}
Olga Russakovsky, Jia Deng, Hao Su, Jonathan Krause, Sanjeev Satheesh, Sean Ma,
  Zhiheng Huang, Andrej Karpathy, Aditya Khosla, Michael Bernstein, et~al.,
\newblock ``Imagenet large scale visual recognition challenge,''
\newblock {\em International Journal of Computer Vision}, vol. 115, no. 3, pp.
  211--252, 2015.

\bibitem{abu2016youtube}
Sami Abu-El-Haija, Nisarg Kothari, Joonseok Lee, Paul Natsev, George Toderici,
  Balakrishnan Varadarajan, and Sudheendra Vijayanarasimhan,
\newblock ``Youtube-8m: A large-scale video classification benchmark,''
\newblock {\em arXiv preprint arXiv:1609.08675}, 2016.

\bibitem{armbrust2010view}
Michael Armbrust, Armando Fox, Rean Griffith, Anthony~D Joseph, Randy Katz,
  Andy Konwinski, Gunho Lee, David Patterson, Ariel Rabkin, Ion Stoica, et~al.,
\newblock ``A view of cloud computing,''
\newblock {\em Communications of the ACM}, vol. 53, no. 4, pp. 50--58, 2010.

\bibitem{hunt2010zookeeper}
Patrick Hunt, Mahadev Konar, Flavio~Paiva Junqueira, and Benjamin Reed,
\newblock ``Zookeeper: Wait-free coordination for internet-scale systems.,''
\newblock in {\em USENIX annual technical conference}. Boston, MA, USA, 2010,
  vol.~8, p.~9.

\end{thebibliography}

\end{document}